%% file: main.tex
\documentclass[%
 reprint,
 amsmath,amssymb,
 aps,
]{revtex4-2}

\usepackage{graphicx}% Include figure files
\usepackage{dcolumn}% Align table columns on decimal point
\usepackage{bm}% bold math
\usepackage{hyperref}

\usepackage{tikz}
\usetikzlibrary{positioning, arrows.meta}

\usepackage{amsmath}
\usepackage{amssymb}
\usepackage{mathtools}
\usepackage{amsthm}
\usepackage{bm}
\theoremstyle{plain}
\newtheorem{theorem}{Theorem}[section]

\theoremstyle{definition}

\newtheorem{assumption}[theorem]{Assumption}
\theoremstyle{remark}

\newcommand{\prob}[1]{\mathbb{P}\left\lbrace#1\right\rbrace}

\newcommand{\avg}[1]{\left\langle #1 \right\rangle}

\usepackage{xcolor}
\definecolor{myblue}{rgb}{0.141, 0.514, 0.596}

\begin{document}

\preprint{APS/123-QED}

\title{Scaling Laws and Representation Learning in Simple Hierarchical Languages: Transformers vs. Convolutional Architectures}
%\title{Manuscript Title:\\with Forced Linebreak}% Force line breaks with \\

\author{Francesco Cagnetta}
\email{francesco.cagnetta@sissa.it}
\affiliation{Scuola Internazionale Superiore di Studi Avanzati (SISSA), Via Bonomea 265, 34136 Trieste, Italy}
\author{Alessandro Favero}
\email{alessandro.favero@epfl.ch}
\affiliation{Institute of Physics, École polytechnique fédérale de Lausanne (EPFL), Lausanne, Switzerland}
\author{Antonio Sclocchi}
\email{a.sclocchi@ucl.ac.uk}
\affiliation{Gatsby Computational Neuroscience Unit, University College London, London, United Kingdom}
\author{Matthieu Wyart}
\email{mwyart1@jh.edu}
\affiliation{Department of Physics \& Astronomy, Johns Hopkins University, Baltimore, MD, U.S.}
\affiliation{on leave from Institute of Physics, École polytechnique fédérale de Lausanne (EPFL), Lausanne, Switzerland}

\date{\today}% It is always \today, today,

\begin{abstract}
How do neural language models acquire a language’s structure when trained for next-token prediction? We address this question by deriving theoretical scaling laws for neural network performance on synthetic datasets generated by the Random Hierarchy Model (RHM)---an ensemble of probabilistic context-free grammars designed to capture the hierarchical structure of natural language while remaining analytically tractable. Previously, we developed a theory of representation learning based on data correlations that explains how deep learning models capture the hierarchical structure of the data sequentially, one layer at a time. Here, we extend our theoretical framework to account for architectural differences. In particular, we predict and empirically validate that convolutional networks, whose structure aligns with that of the generative process through locality and weight sharing, enjoy a faster scaling of performance compared to transformer models, which rely on global self-attention mechanisms. This finding clarifies the architectural biases underlying neural scaling laws and highlights how representation learning is shaped by the interaction between model architecture and the statistical properties of data.
% \begin{description}
% Secondary publications and information retrieval purposes.
% You may use the \texttt{description} environment to structure your abstract;
% use the optional argument of the \verb+\item+ command to give the category of each item. 
% \end{description}
\end{abstract}

\maketitle

%\tableofcontents

\section{Introduction}\label{sec:intro}

\looseness=-1 The exceptional success of deep learning methods in vision, language, and other domains has prompted an urgent search for a theory that can explain their effectiveness. Despite the general absence of a fundamental understanding, practitioners have identified several empirical regularities that any such theory should ultimately aim to predict. Among these, the most prominent are \emph{Neural Scaling Laws} \cite{hestness2017deep, kaplan2020scaling, henighan2020scaling, hoffmann2022training}, which describe power-law relationships between model performance and resource variables such as dataset size, model parameters, and compute budget. On the one hand, these laws have proven remarkably consistent across domains and architectures and now serve as practical heuristics for guiding resource allocation in large-scale machine learning. On the other hand, their ubiquity suggests the presence of fundamental statistical phenomena underlying the behaviour of modern deep learning methods---much like the scaling relations observed near critical points in statistical physics \cite{fisher1967theory}.

If these statistical phenomena are truly universal, then they must reside in the data itself. After all, a learner can generalise only if the data exhibits patterns that are discoverable from finite examples, i.e., learning is impossible if there is nothing to learn. While early theoretical studies often assumed overly simplistic data models---with inputs being independent or exhibiting shallow statistical dependencies, and tasks corresponding to low-dimensional or extremely smooth functions---these assumptions fail to capture the richness of real data. Natural data such as language and images exhibit structure in the form of long-range dependencies and \emph{compositionality}, where complex features are constructed from simpler parts according to systematic rules. Recent work has introduced hierarchically compositional models of data that better reflect the kind of structure deep learning methods exploit, while retaining some degree of analytical tractability~\cite{malach2018provably, malach2020implications, cagnetta2024deep, tomasini2024how, cagnetta2024towards, garnier2024transformers, sclocchi2025phase, sclocchi2025probing, favero2025compositional}. In this work, we build on this perspective by studying how different architectures learn hierarchically compositional data, revealing the impact of architectural differences on the scaling of performance.

Specifically, we focus on synthetic datasets generated by the \emph{Random Hierarchy Model} (RHM)—an ensemble of hierarchically compositional generative processes corresponding to simple context-free grammars~\cite{chomsky1956three}, introduced in~\cite{cagnetta2024deep} to study the role of depth in supervised learning tasks. The learning setup is a \emph{last-token prediction} task, where a deep neural network is tasked with predicting the final token of a sequence of $d$, given the first $d\,{-}\,1$ ones. In~\cite{cagnetta2024towards}, we showed that deep networks use correlations between the tokens to infer the hierarchical structure of the data, and derived the resulting scaling of the test error with the number of training data. The present paper extends this correlation-based scaling analysis to incorporate architectural differences. In particular, we predict that translational invariant architectures such as convolutional networks have access to stronger correlations, therefore enjoy a faster improvement in performance over training than transformers. We empirically validate these theoretical predictions by examining the learning dynamics of deep convolutional networks and transformers trained with standard gradient-based algorithms on our synthetic dataset. Our analysis reveals how fundamental architectural biases affect neural scaling laws.

The remainder of the paper is organised as follows:
\begin{itemize}
    \item In~\autoref{sec:related}, we review related works, including theories of neural scaling laws based on simpler data structures, theories of representation learning based on correlations, and empirical and theoretical works using several classes of formal languages to understand the behaviour of neural language models.
    \item In~\autoref{sec:rhm}, we introduce the Random Hierarchy Model (RHM), an ensemble of synthetic datasets based on probabilistic context-free grammars. The specific assumptions made on the data-generating process render the dataset analytically tractable, enabling the explicit computation of data statistics and their ensemble averages.
    \item In~\autoref{sec:stats}, we characterise the fundamental statistical properties of RHM data, focusing on token correlations and their relationship with the hidden hierarchical structure. 
    \item In~\autoref{sec:theory}, we develop a theoretical framework for learning RHM data in the last-token prediction setting. This framework leverages the link between observable correlations and the hidden tree structure, allowing us to compute sample complexities for inferring hidden variables from a finite dataset of example sentences. 
    \item In~\autoref{sec:architectures}, we describe the neural network architectures studied in this work: convolutional networks, which implement local connectivity and weight sharing, and transformers, which use global self-attention mechanisms to capture long-range dependencies.
    \item In~\autoref{sec:curves}, we compare our theory with the empirical learning curves of deep neural networks. This is our main result, highlighting systematic differences in the scaling behaviour of the test loss between convolutional networks and transformers.
    \item In~\autoref{sec:dynamics}, we qualitatively validate our theory of representation learning by studying how the internal representations of deep networks evolve during training.
    \item In~\autoref{sec:lcn}, we extend the analysis of the previous two sections to locally connected networks, which retain the local receptive field of convolutional networks but without weight sharing, to isolate the role of architectural priors in learning hierarchical data.
    \item Finally, ~\autoref{sec:conclusions} summarises our main finding about the impact on architectural differences on neural scaling law and discusses implications and possible future directions.
\end{itemize}

\section{Related Works}\label{sec:related}

\textbf{Theories of neural scaling laws.} Power-law learning curves can arise in models that memorise training data drawn from a Zipfian (i.e., power-law) distribution~\cite{hutter2021learning,michaud2023quantization}, but such models fail to account for generalisation beyond the training set. Alternatively, power laws emerge in kernel methods when the target function has a power-law spectrum in the kernel eigenbasis~\cite{caponnetto2007optimal}, a mechanism that underpins many theoretical analyses of scaling in neural networks~\cite{spigler2020asymptotic,bordelon2020spectrum,bahri2021explaining,favero2021locality,cui2021generalization,maloney2022solvable,cagnetta2023can,bordelon2024dynamical}. However, these analyses are restricted to kernel or \textit{lazy training} regimes~\cite{jacot2018neural,chizat2019lazy} and do not capture the representation learning capabilities of modern deep networks. A number of works have attempted to go beyond this limitation in simplified models~\cite{paccolat2021geometric, ba2022high,bietti2022learning,dandi2023two,bordelon2025how}, showing that feature learning can modify scaling exponents and improve generalisation.\\

\textbf{Correlation-based representation learning.} Hierarchical generative models on trees, initially developed for phylogeny \cite{mossel2016deep}, have become a powerful tool in understanding both supervised~\cite{malach2018provably, malach2020implications,cagnetta2024deep,tomasini2024how} and self-supervised~\cite{cagnetta2024deep,mei2024u,sclocchi2025phase,favero2025compositional,garnier2024transformers} learning algorithms. In supervised settings, \cite{malach2018provably} introduced a sequential clustering algorithm that highlights the importance of correlations between the input features and the labels for learning. \cite{cagnetta2024deep} introduced the RHM as a framework to show how correlations emerge from the generative model and can be used by deep networks to represent the latent hierarchical structure. In self-supervised settings, several works~\cite{cagnetta2024deep,garnier2024transformers,favero2025compositional} have shown that neural networks progressively learn longer-range correlations, corresponding to deeper layers of the data hierarchy, throughout the course of training. This staged process aligns with the so-called \textit{distributional simplicity bias}, where neural networks sequentially learn statistics of increasing order during training~\cite{refinetti2023neural,bardone2024sliding,rende2024distributional}.\\

\textbf{Learning formal languages.} There is a growing number of studies using generative models from theoretical linguistics to understand the capabilities of large language models, including $n$-grams~\cite{svete2024transformers, nguyen2024understanding, svete2024can}, and regular~\cite{borenstein2024languages, shai2024transformers} and context-free grammars~\cite{allen2023physics, zhao2023transformers}. All these works concern either expressivity or the interpretability of the representations of trained transformers. \cite{zhao2023transformers}, in particular, showed that the operations performed by BERT-like transformers resemble well-known algorithms for grammatical inference, and proved that, for PCFG data, these algorithms are optimal solutions of the masked language modelling objective. However, when the training data is compatible with both a PCFG and a non-hierarchical generative model, neither recurrent language models~\cite{mccoy2020syntax} nor transformer~\cite{ahuja2024learning} consistently prefer the hierarchical explanation. In addition, none of these works study the learning process and the sample complexity.

\section{Context-Free Grammars and the Random Hierarchy Model}\label{sec:rhm}

In this section, we introduce the notion of context-free grammar and our model of hierarchically compositional data.
\begin{figure*}[t]
  \centering

  \begin{minipage}[t]{0.48\textwidth}
    \centering
    \input{parse_tree.tex}
  \end{minipage}
  \hfill
  \begin{minipage}[t]{0.48\textwidth}
    \centering
    \input{rhm_tree}
  \end{minipage}

  \caption{Comparison of structural hierarchies in \textbf{(left)} natural language and \textbf{(right)} an instance of the Random Hierarchy Model (RHM).}
  \label{fig:tree-sketches}
\end{figure*}

\subsection{Context-free grammars and compositional structure}\label{ssec:cgf}

Many real-world data modalities---such as natural language, visual scenes, and even code---exhibit a hierarchical and compositional structure. \emph{Context-Free Grammars} (CGFs) from theoretical linguistics~\cite{chomsky1956three, rozenberg1997handbook} offer a natural formalism to describe such structures\cite{pullum1982natural,joshi1985tree,manning1999foundations,zhu2006stochastic}. A CFG consists of a set of \emph{production rules} for the recursive expansion of abstract hidden (\emph{nonterminal}) symbols into sequences of other nonterminal and observable (\emph{terminal}) symbols. The recursive nature of the rules induces a tree-like structure, called \emph{derivation}---see~\autoref{fig:tree-sketches}, left panel, for an illustration. The leaves of the tree form the observable data, whereas the internal nodes represent the hidden hierarchical structure. The arrows connecting hidden to other hidden or observable nodes represent production rules. 

Each hidden symbol can typically expand into several distinct sequences, each described by a separate production rule. By assigning probabilities to each production rule, a CFG becomes a \emph{Probabilistic Context-Free Grammar} (PCFG), defining a probability distribution over sequences of symbols. Samples from a PCFG are generated by recursively applying production rules until a sequence of terminal symbols is obtained. Compared to simpler generative models such as Markov processes and regular grammars, PCFGs can model complex dependencies spanning unbounded distances~\cite{ebeling1994entropy,lin2017critical}, making them particularly suited for modelling structured real-world data. However, these models are not analytically tractable in general. To address this issue, we consider a simplified yet expressive class of PCFGs: the Random Hierarchy Model (RHM).

\subsection{The Random Hierarchy Model}

The RHM is an ensemble of PCFGs, where, as is customary in statistical mechanics, the production rules are drawn \emph{uniformly at random}, compatibly with the following set of constraints.
\begin{itemize}
\item[C1.] \textbf{(Fixed tree topology)} All grammars share a common tree structure: a fixed regular tree of arity $s$ and depth $L$. An example having $s\,{=}\,2$ and $L\,{=}\,3$ is displayed in~\autoref{fig:tree-sketches}, right, where hidden variables are represented as variable nodes (empty circles) and we used factor nodes (full squares) to represent production rules. This tree serves as the shared backbone for all the generated data, which consists of sequences of $d\,{=}\,s^L$ symbols. Because of the fixed topology constraint, exact inference on data generated by the RHM can be performed using the Belief Propagation (BP) algorithm~\cite{mezard2009information}. BP enables the efficient computation of posterior distributions over hidden symbols given observed data, as well as marginal distributions over observed symbols. In particular, it yields closed-form expressions for data statistics---such as symbol frequencies and pairwise correlations---for any given grammar in the ensemble. Moreover, since the grammars are sampled randomly, these quantities can be further averaged over the ensemble, allowing us to calculate expected values and variances of relevant observables analytically.
\item[C2.] \textbf{(Unambiguity)} Production rules are chosen so that no two distinct hidden symbols are allowed to generate the same sequence of children. This constraint ensures that the observable datum uniquely determines the hidden structure of its derivation. Without some form of unambiguity constraint, a PCFG with randomly sampled production rules would typically generate sequences with weak or trivial correlations: the observable tokens would become nearly independent, losing any statistical imprint of the underlying hierarchical structure~\cite{hsu2012identifiability, degiuli2019random}.
\item[C3.] \textbf{(Vocabulary size $v$)} Hidden symbols are split into $L$ vocabularies $\mathcal{V}_\ell$, with $\ell\,{=}\,0,\dots,L-1$ and $\mathcal{V}_L\,{\equiv}\mathcal{V}$ denoting the vocabulary of observable symbols. All vocabularies have the same cardinality $v$.
\item[C4.] \textbf{($m$ production rules per symbol)} Each hidden symbol is associated with $m$ distinct and equiprobable production rules that yield symbols of the next level. Due to unambiguity, $mv\,{\leq} v^s$, i.e. $f\,{\coloneq}\, (m/v^{s-1})\,{\leq}\,1$.
\end{itemize}
Given an instance of the RHM, the data-generating process works as follows:
\begin{itemize}
\item[i)] Sample a root symbol $\mu^{(0)}\in \mathcal{V}_0$ according to the uniform probability $1/v$ (e.g. $a$ in~\autoref{fig:tree-sketches}, right);
\item[ii)] Sample one of the $m$ production rules expanding $\mu^{(0)}$ into $s$ symbols from $\mathcal{V}_1$,
\begin{equation}\label{eq:production-rule}
\mu^{(0)} \rightarrow \mu^{(1)}_1,\dots,\mu^{(1)}_s,
\end{equation} (e.g. $a\rightarrow b,c$ in~\autoref{fig:tree-sketches}, right);
\item[iii)] replace $\mu^{(0)}$ with the right-hand side of~\autoref{eq:production-rule} to obtain the next-level representation;
\item[iv)] Repeat steps ii) and iii) above for each element of the new representation and iterate $L$ times. 
\end{itemize}
This process defines a probability distribution over sequences in $\left(\mathcal{V}\right)^d$,
\begin{equation}\label{eq:data-prob}
P_{\bm{X}}(\bm{x})=\prob{X_1=x_1,\dots,X_{d}=x_{d}},
\end{equation}
where the probability of each datum $\bm{x}$ is the product of probabilities for all production rules involved in its generation. Following standard Language Modelling jargon, we refer to the components of $\bm{x}$ as \emph{tokens}.

\section{Statistics of RHM data}\label{sec:stats}

Before turning to the question of learnability, we first examine the statistical structure of the sequences generated by the RHM. In particular, we study the correlations between tokens that are induced by the underlying hierarchical generative process. These correlations reflect the hidden tree structure shared by data and form the primary observable cues that a learning algorithm can exploit to infer such structure.

We define the correlation between the $i$-th and $j$-th token of the sequence via the $v\times v$ co-occurrence matrix $C(X_i,X_j)$, having elements
\begin{align}\label{eq:corr-def}
C(X_i,X_j)_{\mu,\nu}\coloneq& \prob{X_i=\mu,X_j=\nu}\nonumber\\&-\prob{X_i=\mu}\prob{X_j=\nu}.
\end{align}
The tree-like structure gives a simple recipe for computing all the marginals appearing in~\autoref{eq:corr-def}: draw the path that connects the root of the tree to the tokens $X_i$ and $X_j$, multiply the root prior probability $p(\mu^{(0)})$ by one transition probability per every factor node traversed by the path, then sum over all the hidden variables along the path. For instance, the one-token marginal for the first token of the sequence in~\autoref{fig:tree-sketches}, right, is given by
\begin{align}\label{eq:1marginal-example}
\prob{X_1=\mu} = &\sum_{\mu^{(2)}_1} p(\mu|\mu^{(2)}_1) \sum_{\mu^{(1)}_1} p(\mu^{(2)}_1|\mu^{(1)}_1) \times \nonumber\\ &\sum_{\mu^{(0)}} p(\mu^{(1)}_1|\mu^{(0)}) p(\mu^{(0)}),
\end{align}
where $\mu^{(0)}$ denotes the root symbol, $\mu^{(1)}_1$ the first level-$1$ symbol,  $\mu^{(2)}_1$ the first level $2$ symbol, and $p(\mu^{(\ell)}|\mu^{(\ell-1)})$ the transition probability associated with the corresponding production rule. The definition in~\autoref{eq:corr-def} can be directly extended from observable tokens to hidden variables and tuples thereof.

The RHM ensemble provides specific statistics for the transition probabilities associated with the production rules. Using these statistics, we can obtain analytical expressions for the ensemble averages and variances of various correlation functions, as shown in~\cite{cagnetta2024deep} for the token-root and tuple-root correlations, in~\cite{cagnetta2024towards} for token-token and tuple-token correlations, and in~\cite{favero2025compositional} for the correlations between observable tokens and tuples of hidden variables. All correlations have zero mean over RHM realisations, whereas the variance is given by the following simple formula. Given two hidden or observable variables $(X_i,X_j)$ and their tree distance $d_{\text{tree}}(X_i,X_j)$, then, if $X_i$ is an ancestor of $X_j$ or vice-versa~\cite{cagnetta2024deep},
\begin{align}\label{eq:corr-rhm-anal-ancestors}
\avg{\left(C(X_i,X_j)_{\mu,\nu}\right)^2}_{\text{RHM}} \simeq \frac{1}{v^2}\frac{1}{vm^{d_{\text{tree}}(X_i,X_j)}},
\end{align}
otherwise, if $X_i$ and $X_j$ are not on the same branch~\cite{cagnetta2024towards},
\begin{align}\label{eq:corr-rhm-anal}
\avg{\left(C(X_i,X_j)_{\mu,\nu}\right)^2}_{\text{RHM}} \simeq \frac{1}{v^2}\frac{(1-f)}{vm^{d_{\text{tree}}(X_i,X_j)}},
\end{align}
where the approximation $\simeq$ is exact in the limit of large $m$ and $v$. This result represents a generic feature of RHM data: the correlations between any two symbols decay exponentially with the tree distance, i.e., the number of links in the shortest path connecting said two symbols. 

\subsection{Correlations and hidden variables}\label{ssec:corr-latent}

Due to the context-free structure, correlations involving the entire span of some hidden variable are representative of the value of such a variable. Consider, for instance, the joint probability $\prob{X_1=\mu}$ of the first $2$ tokens for sentences generated by a RHM with $s\,{=}\,2$ and $L\,{=}\,3$ (as in the tree of~\autoref{fig:tree-sketches}, right).
\begin{align}\label{eq:smarginal-example}
\prob{X_1=\mu_1,X_2\,{=}\,\mu_2} = \sum_{\mu^{(2)}_1} p(\mu_1,\mu_2|\mu^{(2)}_1) \times\nonumber\\ \sum_{\mu^{(1)}_1} p(\mu^{(2)}_1|\mu^{(1)}_1)\sum_{\mu^{(0)}} p(\mu^{(1)}_1|\mu^{(0)}) p(\mu^{(0)}).
\end{align}
Due to the unambiguity constraint C2, there is only one value of the latent variable $\mu_1^{(2)}$ compatible with $\mu_1,\mu_2$. Hence, the right-hand side of~\autoref{eq:smarginal-example} simplifies, revealing that the joint probability $\prob{X_1=\mu_1,X_2\,{=}\,\mu_2}$ depends on $\mu_1^{(2)}$ rather than $(\mu_1,\mu_2)$. In other words, given the latent variable $\mu_1^{(2)}$, this probability is conditionally independent of $(\mu_1,\mu_2)$. This property extends to all joint probabilities that include the entire span of some hidden variable in their argument, and implies that the statistics of the span can be used to infer the corresponding hidden variable.

\section{A Theory of Learning the RHM}\label{sec:theory}

We now turn to the question of how the structured data generated by the Random Hierarchy Model can be learned. In this section, we develop a theoretical framework for analysing the learnability of the RHM by focusing on a simplified prediction task, which is defined precisely in~\autoref{ssec:task}. Our approach centres on understanding how a learner can exploit the statistical dependencies induced by the generative process, described in~\autoref{sec:stats}, to recover information about the hidden hierarchical structure. This setting provides a tractable and analytically grounded perspective on the general problem of learning compositional data.

\subsection{The Last-token prediction task}\label{ssec:task}

Several Language Modelling techniques have been developed to provide approximations of~\autoref{eq:data-prob}. For example, Masked Language Modelling randomly replaces a fraction of tokens with a special symbol $x_{\text{mask}}$, and the learner is tasked with inferring their original value from the context~\cite{devlin2019bert}. In contrast, Autoregressive language models sequentially predict each token based on all preceding tokens in the sequence~\cite{radford2018improving}. Here, following~\cite{cagnetta2024towards}, we consider a simplified setup where the learner is tasked with predicting the last token of the sequence from the remaining tokens, or context. Due to the focus on the end of the sequence, we will follow the notation of~\autoref{fig:tree-sketches}, right, and use the subscript $-1$ for the last token of the sequence, $-2$ for the penultimate, and so on. In this notation, the learner aims to approximate the conditional probability
\begin{align}\label{eq:conditional}
P(X_{-1} \mid X_{-2}=x_{-2},\dots,X_{-d}=x_{-d}),
\end{align}
from a training set of $P$ samples of valid RHM sequences. We denote the learner's approximation with $\hat{P}(X_{-1}|x_{-(2:d)})$. The quality of the approximation is measured via the \emph{test cross-entropy loss},
\begin{align}\label{eq:cross-ent-loss}
  \mathcal{L}(\hat{P}) = -\mathbb{E}_{\bm{x}\sim P_{\bm{X}}}\left[\log{\hat{P}(X_{-1}=x_{-1}|x_{-(2:d)})}\right].
\end{align}

\subsection{Flat vs. hierarchical approximation: the role of $n$-grams}\label{ssec:ngrams}

A standard strategy in classical language modelling is to approximate the conditional distribution of the next token based on fixed-length contexts. In the last-token prediction task, this leads to an \emph{n-gram approximation} of the form
\begin{equation}
P(X_{-1} \mid X_{-2}, \dots, X_{-d}) \approx P(X_{-1} \mid X_{-2}, \dots, X_{-n}),
\end{equation}
where the model estimates the probability of $X_{-1}$ given the most recent $n\,{-}\,1$ tokens. Being agnostic of the underlying structure, this approximation requires sampling all length-$n$ contexts. As the number of these contexts grows exponentially with $n$, this approximation is not data efficient.

In contrast, taking the hidden context-free structure into account yields a much more efficient strategy. Indeed, as shown in~\autoref{ssec:corr-latent}, given a hidden variable, its span is conditionally independent of the rest of the sequence. As a result, the content of a length-$n$ context is determined by fewer than $n$ variables: using these variables would dramatically reduce the number of contexts to sample. In the specific case of last-token prediction, the variables contributing to the prediction of $X_{-1}$ include:
\begin{itemize}
\item The $(s\,{-}\,1)$ `siblings' $(X_{-s},\dots,X_{-2})$, generated with $X_{-1}$ by the last level-${L}$ hidden variable $\mu^{(L)}_{-1}$;
\item The $(s\,{-}\,1)$ siblings of $\mu^{(L)}_{-1}$, and of all the other ancestors $\mu^{(\ell)}_{-1}$ of $X_{-1}$;
\item The root symbol $\mu^{(0)}$.
\end{itemize}
The total count sums up to $(s\,{-}\,1)L\,{+}1$, exponentially smaller than the full input length $d\,{=}\,s^L$.

The contrast between the flat and hierarchical $n$-gram approximations reveals the possibility of learning the conditional last-token probability efficiently by leveraging the hierarchical structure of the data. The efficient strategy requires knowledge of the hidden structure of the data, which is not explicitly provided to the learner. However, as we will demonstrate in the next subsection, an ideal learner could use observable statistics---specifically, token-token correlations---to reconstruct the hidden hierarchical structure of the data.

\subsection{Reconstruction of the hidden variables via correlations}\label{ssec:reconstruction}

To access the hierarchical $n$-gram approximation, a learner requires access to the hidden variables, which are not explicitly provided in the training data. However, as shown in~\autoref{ssec:corr-latent}, the joint statistics of spans of tokens generated by the same hidden variable are representative of the hidden variable itself. As a result, a learner could use these statistics to cluster all the spans corresponding to the same hidden variable, i.e. infer this variable. The only obstacle to inference is the finite number of training examples, implying that the learner has access only to noisy measurements of such statistics.

\subsubsection{Reconsturction of depth-$1$ hidden variables}

For concreteness, let us consider the correlations between the last token $X_{-1}$ and one of the entire $s$-tuples in the context, $\bm{X}_{-t}\,\coloneq\,(X_{-ts},\dots,X_{-(t-1)s+1)})$ with $t\,{=}\,2,\dots,s^{L-1}$. These correlations are given by the following $(vm)\times v$
\begin{align}\label{eq:corr-tuple-token}
C(\bm{X}_{-t},X_{-1})_{\bm{\mu},\nu}\coloneq& \prob{\bm{X}_{-t}=\bm{\mu},X_{-1}=\nu}\nonumber\\&-\prob{\bm{X}_{-t}=\bm{\mu}}\prob{X_{-1}=\nu}.
\end{align}
According to~\autoref{sec:stats}, entries corresponding to the $m$ distinct $\bm{\mu}$'s generated by the same latent variable $\mu^{(L)}_{-t}$ are all equal to each other. However, the noise induced by the finite training set might hinder the inference of $\mu^{(L)}_{-t}$. For a large training set size $P$, such noise is equivalent to an additive Gaussian noise with zero mean and variance~\cite{cagnetta2024towards}
\begin{align}\label{eq:sampling-noise}
\sigma^2_P = \frac{1}{(vm)v P}.
\end{align}
If this variance is larger than the difference between correlations associated with distinct hidden variables, then inference of the hidden variable is impossible.

We estimate the difference between correlations corresponding to different hidden variables with the variance of $C(\bm{X}_{-t},X_{-1})_{\bm{\mu},\nu}$ over realisations of the RHM~\cite{cagnetta2024towards},
\begin{align}\label{eq:corr-rhm-tuple-token}
&\avg{\left(C(\bm{X}_{-t},X_{-1})_{\bm{\mu},\nu}\right)^2}_{\text{RHM}} \nonumber \\&= \avg{\frac{1}{m}\left(C(X^{(L)}_{-t},X_{-1})_{\bm{\mu},\nu}\right)^2}_{\text{RHM}} \nonumber\\
&\simeq \frac{1}{(vm)v}\frac{(1-f)}{vm^{d_{\text{tree}}(X^{(L)}_{-t},X_{-1})}}.
\end{align}
Comparing~\autoref{eq:sampling-noise} with~\autoref{eq:corr-rhm-tuple-token}, we get the following condition for the inference of the depth-$1$ hidden variable $\mu^{(L)}_{-t}$,
\begin{align}
P\,{\gg} (1-f)^{-1} v m^{d_{\text{tree}}(X^{(L)}_{-t},X_{-1})}.
\end{align}
Due to the fixed tree topology assumption C1, this condition corresponds to $L\,{-}\,1$ different sample complexities for the inference of the hidden variables $\mu^{(L)}_{-t}$ with $t\,{<}\,s^{\ell}$ with $\ell\,{=}\,1,\dots,L-1$,
\begin{align}\label{eq:sc-general}
P_\ell = (1-f)^{-1} v m^{2\ell+1}.
\end{align}

\subsubsection{Reconstruction of deeper hidden variables}

The same ideas apply for the inference of the next layer of hidden variables, i.e., depth-$2$ or level-$(L\,{-}\,1)$ variables, via correlations between $X_{-1}$ and $s$-tuples of level-$L$ variables. These tuples have a smaller tree distance from $X_{-1}$ than tuples of input tokens, thus the corresponding sample complexities are lower than those required by~\autoref{eq:sc-general}. Hence, when $P\,{\gg}\,P_{\ell}$, all the latent variables spanning the context going $X_{-s^{\ell+1}}$ to $X_{-(s+1)}$ can be inferred. We can formalise this concept via the following assumption:
\begin{assumption}
A learner discovers the hidden structure up to depth $\ell$ when the correlations between the last token $X_{-1}$ and the $s$-tuples of observable tokens $\bm{X}_{-t}$ at tree distance $d_{\text{tree}}(\bm{X}_{-t},X_{-1})\,{=}\,2\ell+1$ become detectable from the training data.
\end{assumption}

\subsubsection{Improved sample complexities due to translation invariance}\label{sssec:reconstruction-cnn}

In the RHM, the same set of production rules applies to all variables at a given level, independently of the position. Therefore, a learner could infer all level-$L$ latents as soon as it's able to infer $\mu^{^{(L)}}_{-2}$, which is the easiest to infer as it has the strongest correlation with $X_{-1}$. If this were the case, we could assume the learner can access all the level-$L$ hidden variables with $P\,{\gg}\,P_1$ data. This would improve the sample complexities for the inference of deeper hidden variables, as the learner could correlate the last token directly with tuples of hidden variables, having stronger correlations than the tuples of observable tokens. The corresponding condition for the sample complexities required to infer all latents of depth $\ell$ and smaller is~\cite{favero2025compositional}
\begin{align}\label{eq:sc-improved}
 P \gg \bar{P}_{\ell}\coloneq(1-f)^{-1} v m^{\ell+2}.
\end{align}
As we demonstrate empirically in~\autoref{sec:curves}, the performance scaling of Convolutional Neural Networks is compatible with this improved strategy, whereas the scaling of Transformers follows~\autoref{eq:sc-general}.

\section{Architectures}\label{sec:architectures}

In this section, we describe the different architectures considered in this paper.

\subsection{Convolutional Neural Networks (CNNs)}

Convolutional Neural Networks (CNNs) are a class of architectures originally developed for image processing~\cite{lecun1989backpropagation}, where they exploit local connectivity and translation invariance to efficiently model spatial patterns. Beyond images, CNNs have also been successfully applied to sequential data, such as text~\cite{kalchbrenner2014convolutional,kim2014convolutional}, by adapting the operations to one-dimensional inputs. In this work, we focus on one-dimensional CNNs, which are well-suited for sequence modelling. More specifically, our CNN architectures are designed to match the hierarchical structure of the RHM. Each network consists of \(L\) hidden layers, with each layer corresponding to one level of the underlying tree. Filters operate without overlap: the filter size and stride are both set to the RHM branching factor \(s\), and no padding is used. 

CNNs receive as input a sequence of one-hot encodings of the $d$ input features: each token $x_i$ is a $v$-dimensional vector, whose only non-vanishing component coincides with the vocabulary entry. The $(d\,{-}\,1)$ observable tokens are whitened, so as to have $\sum_{\mu=1}^v x_{i,\mu} \,{=}\,0$, $\sum_{\mu=1}^v (x_{i,\mu})^2 \,{=}\,1$ for all $i$'s. The last token is replaced with the dummy token $x_{\mu}\,{=}\,v^{-1/2}$. The activations \(h_{\ell}\) at layer \(\ell\) are defined recursively as
\begin{equation}
h_{\ell,i} = \sigma\left(W^{(\ell)} h_{\ell-1,s i : s(i+1)-1} + b^{(\ell)}\right),
\end{equation}
where \(W^{(\ell)}\) and \(b^{(\ell)}\) denote the weight matrix and bias at layer \(\ell\), \(\sigma(\cdot)\) is a pointwise nonlinearity (e.g., ReLU), and \(h_{\ell-1,s i : s(i+1)-1}\) denotes the \(s\)-dimensional slice of the previous layer’s activations corresponding to the \(i\)-th output at level \(\ell\). The weights $W^{(\ell)}$ are order-$3$ tensor, $W^{(\ell)}\in \mathbb{R}^{w_{\ell}\times w_{\ell-1}\times s}$, with $w_0\,{=}\,v$ and $w_{\ell}\,{=}\,w$ for $\ell\,{\geq}\,1$. The output is obtained by applying a $v\times w$ linear readout $a$ on the last hidden layer,
\begin{align}
f_{\text{CNN}}(x_{-(2:d)}) = w^{-1} a \cdot h_{L}(x_{-(2:d)}).
\end{align}

We consider the infinite-width limit $w\to\infty$, which in our setting is realised around $w\,{=}\,512$, and we verified empirically that increasing the width further does not affect the results presented in the paper. The parameters are trained with Stochastic Gradient Descent (SGD) on the cross-entropy loss~\autoref{eq:cross-ent-loss}, with learning rate $w$ and batch size $128$. We consider the setting of \emph{online training}, where a new batch of data is generated independently at each training step, so that the total number of training data is linearly proportional to the number of steps.

\subsection{Transformers}

Transformer architectures have emerged as a central paradigm in modern machine learning~\cite{vaswani2017attention}. Their key innovation is the self-attention mechanism, which enables models to aggregate information from different parts of an input sequence, irrespective of their distance, and thus capture long-range dependencies effectively.

We consider a simplified one-dimensional transformer consisting of \( L \) hidden layers. Each layer applies a multi-head self-attention operation over the full sequence, without any additional feedforward operations. Unlike CNNs, transformers preserve the sequence length across layers and can directly model global interactions between all input tokens.

Mathematically, the transformer requires first a projection of the input tokens into an \emph{embedding} space of dimension $d_K$, which we achieve via a learnable $v\times d_K$ matrix acting on the one-hot encodings of the tokens. A random and learnable $d_K$-dimensional sequence, known as \emph{positional encoding}, is added to each token to break the permutation equivariance of the rest of the architecture. The following activations \( h_{\ell,i} \) at layer \( \ell \) and position \( i \) are given by
\begin{equation}
h_{\ell,i} = \mathrm{MHA}^{(\ell)}\left(h_{\ell-1,1}, \dotsc, h_{\ell-1,d}\right)_i,
\end{equation}
where \(\mathrm{MHA}^{(\ell)}\) denotes the multi-head attention operation at layer \(\ell\), and \(h_{0}\) represents the input embeddings. In particular, each MHA layer first computes, for each head \( h\), learned projections of the previous layer outputs:
\(Q^{(\ell,h)}_i {=} W_Q^{(\ell,h)} h_{\ell-1,i},\),
\(K^{(\ell,h)}_i {=} W_K^{(\ell,h)} h_{\ell-1,i}\),
\(V^{(\ell,h)}_i {=} W_V^{(\ell,h)} h_{\ell-1,i}\), where \( W^{(\ell,h)}\)'s are trainable weight matrices. For each head, attention scores \(\alpha^{(\ell,h)}\)'s and head outputs are then computed as
\(\alpha^{(\ell,h)}_{ij} = \mathrm{softmax}_j\left( d_k^{-1/2} \, Q^{(\ell,h)}_i \cdot K^{(\ell,h)}_j \right),\,
\text{head}^{(\ell,h)}_i = \sum_{j=1}^d \alpha^{(\ell,h)}_{ij} V^{(\ell,h)}_j\), with \( d_k \) denoting the dimension of the key vectors. The outputs of all heads are concatenated and projected to produce the layer output:
\begin{equation}
\mathrm{MHA}^{(\ell)}(h_{\ell-1,1}, \dotsc, h_{\ell-1,d})_i = W_O^{(\ell)} \text{multihead}^{(\ell)} ,
\end{equation}
where \( W_O^{(\ell)} \) is learned and \(\text{multihead}^{(\ell)} =  \text{concat}(\text{head}^{(\ell,1)}_i, \cdots, \text{head}^{(\ell,H)}_i)\).
This global operation enables each token to aggregate information from the entire sequence at each layer. However, the absence of built-in locality biases requires the model to learn the hierarchical structure from the data.

The output of the architecture is obtained by applying a $v\times w$ linear readout $a$ on the last token of the last hidden layer,
\begin{align}
f_{\text{MLA}}(x_{-(2:d)}) = w^{-1} a \cdot h_{L,-1}(x_{-(2:d)}).
\end{align}
We set $d_K\,{=}\,512$ and the number of heads to $16$, and we verified empirically that our results are stable to changing the number of heads at fixed $d_K$ and to varying $d_K$ in the range $[256,512]$ with the number of heads fixed to $d_K/32$. The parameters of Transformers are trained with the Adam optimiser---a variant of SGD that adapts the learning rate for each parameter based on its gradient history---on the cross-entropy loss~\autoref{eq:cross-ent-loss}, with learning rate $10^{-3}$ and batch size $128$. As for CNNs, we consider the \emph{online training} setting.

\section{Predicted Scaling Laws vs. Empirical Learning Curves}\label{sec:curves}

In this section, we compare the predictions outlined in~\autoref{sec:theory} with the training dynamics of deep transformers and CNNs. We consider an online training setup, where a new batch of  $B\,{=}\,128$ training data is sampled independently at each training step. As a result, the total number of training data is proportional to the number of training steps.

\subsection{Transformers}

\begin{figure}
    \centering
    \includegraphics[width=\linewidth]{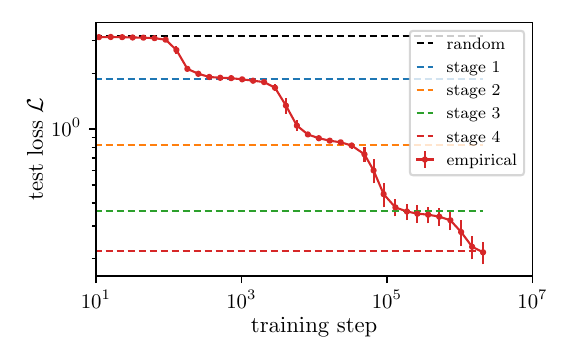}
    \caption{Training dynamics of a depth-$4$ Transformer trained on RHM data with $L=4$, $s=2$, $v=24$, and $m=6$. At each step of training, the test cross-entropy loss is evaluated on a test set of size $2^{15}$ and averaged over $8$ independent realisations of the RHM. This procedure applies to all figures in the paper unless otherwise stated. Colored dashed lines show the averaged $s^{\ell}$-gram losses defined in~\autoref{eq:slgram-losses}: as the number of training steps increases, the Transformer’s performance transitions between successive approximations.}
    \label{fig:mla-example}
\end{figure}

~\autoref{fig:mla-example} illustrates the stagewise dynamics of deep transformers, mirroring the stagewise learning curves observed in~\cite{cagnetta2024towards}. The dashed lines represent the random-choice loss (or the $s^{0}$-gram) and the ensemble-averaged $s^{\ell}$-grams losses,
\begin{align}\label{eq:slgram-losses}
  \mathcal{L}_{\ell} \coloneq \avg{\mathbb{E}_{\bm{x}\sim P_{\bm{X}}}\left[-\log{P(X_{-1}=x_{-1}|x_{-(2:s^{\ell})})}\right]}_{RHM}
\end{align}
with $\ell\,{=}\,1,\dots,L$. While the random-choice loss is simply given by $\log{v}$, the $\mathcal{L}_{\ell}$'s are computed by sampling $64$ independent instances of the RHM with fixed $L$, $s$, $v$ and $m$, obtaining the losses of each instance exactly via Belief Propagation, then averaging over the sampled instances. As the number of training steps increases, the performance jumps from one approximation to the next.

\begin{figure}
    \centering
    \includegraphics[width=\linewidth]{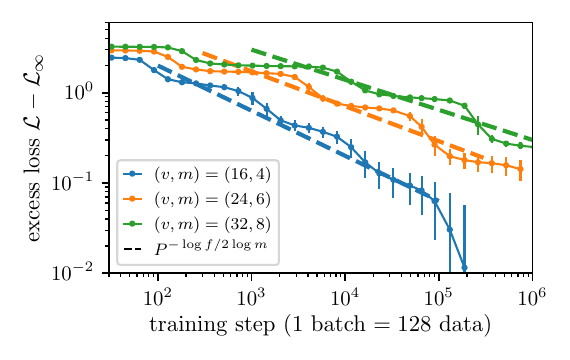}
    \caption{Scaling of the excess test loss, $\mathcal{L} - \mathcal{L}_{\infty}$, as a function of the number of training steps for depth-$4$ Transformers trained on RHMs with shared tree structure ($L=4$, $s=2$) but varying values of $m$ and $v$ (indicated in the legend). Each parameter set is associated with a distinct color. Solid lines show empirical results, while dashed lines represent scaling predictions from~\autoref{eq:scaling-law-general}.}
    \label{fig:mla-scaling}
\end{figure}

The scaling law is obtained by inverting the right-hand side of~\autoref{eq:sc-general} for $\ell$ as a function of $P$, then substituting it into the asymptotic behaviour of $\mathcal{L}_{\ell}$ with $\ell$ found in~\cite{cagnetta2024deep}:
\begin{align}\label{eq:scaling-law-general}
\mathcal{L}_{\ell}-\mathcal{L}_\infty \sim f^{\ell},\quad P_\ell \sim m^{2\ell} \Rightarrow   \mathcal{L}(P)-\mathcal{L}_\infty \sim P^{\frac{\log{f}}{2\log{m}}}.
\end{align}
As shown in~\autoref{fig:mla-scaling}, ~\autoref{eq:scaling-law-general} correctly predicts the decay of the loss of deep transformer towards the limiting value $\mathcal{L}_{\infty}$ for several values of the RHM parameters $m$ and $v$.

\subsection{Hierarchical CNNs}

\begin{figure*}
    \centering
    \includegraphics[width=\linewidth]{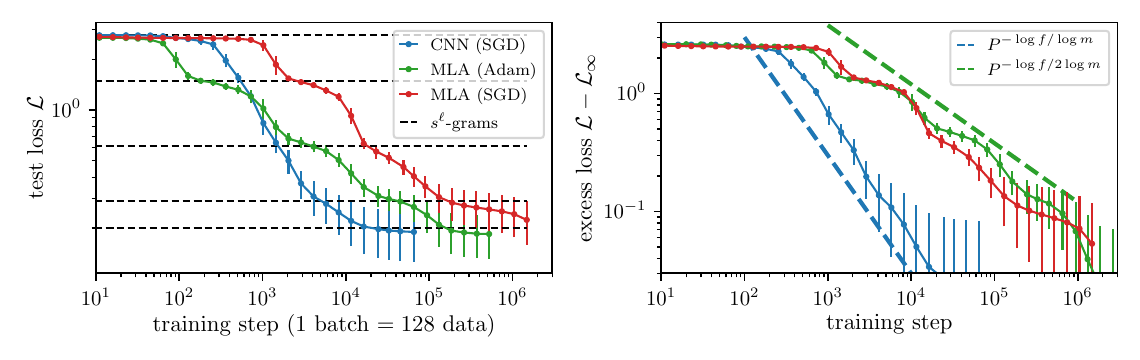}
    \caption{Training dynamics of depth-$4$ models on RHM data with $L\,{=}\,4$, $s\,{=}\,2$, $v\,{=}\,16$, and $m\,{=}\,4$. \textbf{Left:} Test cross-entropy loss as a function of the number of training steps for depth-$4$ Transformers trained with Adam and vanilla SGD, and for depth-$4$ CNNs trained with SGD. Black dashed lines indicate the $s^{\ell}$-gram losses. CNNs exhibit a much faster loss decay compared to Transformers, and a constant delay is observed between the Adam- and SGD-trained Transformers, with Adam reaching lower losses earlier. \textbf{Right:} Scaling of the excess test loss, $\mathcal{L} - \mathcal{L}_{\infty}$, as a function of the number of training steps. The Adam curve is rescaled to overlap with the Transformer SGD curve, highlighting their similar scaling behaviour. By contrast, CNNs display a different scaling behaviour. Both CNN and Transformer scalings are compared with the corresponding theoretical predictions reported in the legend.}
    \label{fig:cnn-vs-mla}
\end{figure*}

~\autoref{fig:cnn-vs-mla} compares dynamics of a deep transformer (green curve) to that of a deep CNN tailored to the structure of the RHM (blue curve), i.e. having $L$ hidden layers, with filter size $s\,{=}\,2$ and stride equal to $s$. The CNN dynamics crosses the $4$ $s^{\ell}$-gram approximation stages sequentially, although the decay is significantly faster than for the transformer. This is highlighted in the right panel, where the asymptotic entropy of the dataset $\mathcal{L}_\infty$ is subtracted. As the CNN is trained with vanilla SGD instead of Adam, we also included the dynamics of a transformer trained with vanilla SGD, to check that the difference in scaling is not due to the difference in the optimiser. While the optimiser causes a shift of the dynamics, the two transformer curves (green and red) display the same decay towards $\mathcal{L}_\infty$ . This is confirmed in the right panel of the figure, where the two curves are made to overlap after rescaling the x-axis of the green curve by a constant factor.

\begin{figure}
    \centering
    \includegraphics[width=\linewidth]{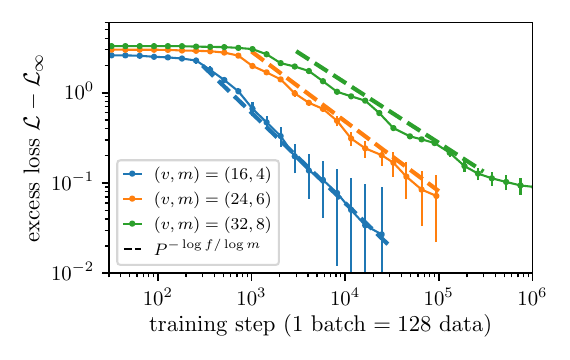}
    \caption{Scaling of the excess test loss, $\mathcal{L} - \mathcal{L}_{\infty}$, as a function of the number of training steps for depth-$4$ CNNs trained on RHMs with shared tree structure ($L=4$, $s=2$) but varying values of $m$ and $v$ (indicated in the legend). Each parameter set is associated with a distinct colour. Solid lines show empirical results, while dashed lines represent scaling predictions from~\autoref{eq:scaling-law-improved}.}
    \label{fig:cnn-scaling}
\end{figure}

As discussed in~\autoref{sssec:reconstruction-cnn}, CNNs could use the weight sharing property of the architecture to exploit stronger correlations than those used by the Transformer, thus enjoying a faster loss decay. To test this hypothesis, we compare the CNN dynamics with the scaling obtained by substituting the relationship between $\ell$ and $P$ from~\autoref{eq:sc-improved} into the asymptotic behaviour of $\mathcal{L}_{\ell}$ with $\ell$:
\begin{align}\label{eq:scaling-law-improved}
\mathcal{L}_{\ell}-\mathcal{L}_\infty \sim f^{\ell},\quad \bar{P}_\ell \sim m^{\ell} \Rightarrow   \mathcal{L}(P)-\mathcal{L}_\infty \sim P^{\frac{\log{f}}{\log{m}}},
\end{align}
to be compared with~\autoref{eq:scaling-law-general} of the previous section. \autoref{fig:cnn-scaling} shows that the loss scaling of deep CNNs agrees with~\autoref{eq:scaling-law-improved} for several values of the RHM parameters $v$ and $m$.

\section{Dynamics of the Hidden Representations}\label{sec:dynamics}

To further understand how models trained on RHM data build hierarchical representations, we study the behaviour of hidden activations under specific input transformations~\cite{cagnetta2024towards}. Given an input sequence $\bm{x}$ and its associated generative tree, we consider two types of transformations:
\begin{itemize}
    \item \textbf{Random variable replacement} $\mathcal{S}_{\ell,i}$: replace the $i$-th hidden variable at level $\ell$ of the tree with another symbol randomly chosen from the vocabulary. This modifies the entire subtree rooted at that variable.
    \item \textbf{Random rule replacement} $\mathcal{R}_{\ell,i}$: resample the production rule emanating from the $i$-th variable at level $\ell$, while keeping the variable itself fixed. This alters the subtree while preserving the identity of the hidden variable.
\end{itemize}

Following~\autoref{sec:architectures}, we denote the hidden representation at layer $\ell$ and position $i$ by $h_{\ell,i}(x)$. For Transformers, $i\,{=}\,-1,\dots,-d$ independently of $\ell$, whereas, for hierarchical CNNs, $i\,{=}\,-1,\dots,-s^{L-\ell}$. To probe their structure, we consider the standardised representations
\begin{equation}
\hat{h}_{\ell,i}(\bm{x}) = \frac{h_{\ell,i}(\bm{x}) - \mathbb{E}_{\bm{x}}[h_{\ell,i}(\bm{x})]}{\sqrt{\mathrm{Var}_{\bm{x}}[h_{\ell,i}(\bm{x})]}} ,
\end{equation}
where the mean and variance are taken over the RHM data. The invariance of a representation to a transformation $\mathcal{T}$ is quantified via the cosine similarity between the original and transformed standardised representations:
\begin{equation}
    q_{\mathcal{T}}(h_{\ell,i}) = \mathbb{E}_{\bm{x}}\left[ \hat{h}_{\ell,i}(\bm{x}) \cdot \hat{h}_{\ell,i}(\mathcal{T}\bm{x}) \right].
\end{equation}
A drop in $q_{\mathcal{T}}(h)$ indicates that the representation is sensitive to the transformation. In particular, invariance to $\mathcal{R}_{\ell,i}$ but sensitivity to $\mathcal{S}_{\ell,i}$ signals that the representation encodes the hidden symbol itself but not the details of its generated subtree.

% \subsubsection{Transformers}
\begin{figure*}
    \centering
    \includegraphics[width=\linewidth]{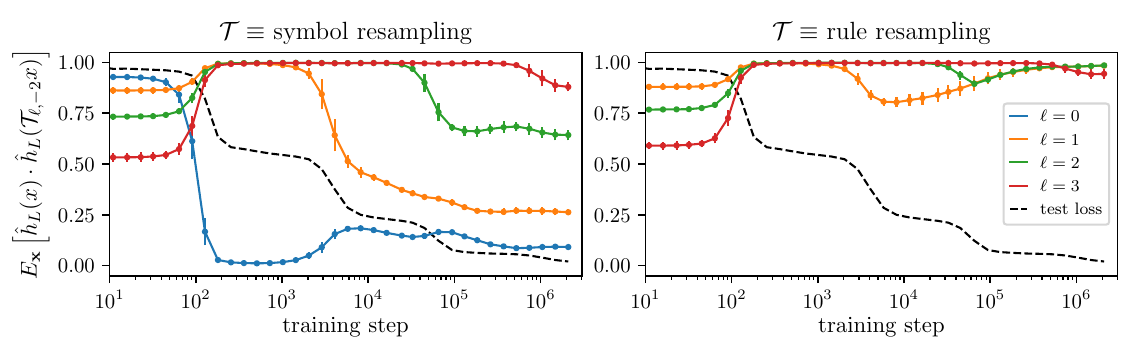}
    \caption{Dynamics of hidden representations of depth-$4$ Transformers trained on RHM data with $L\,{=}\,4$, $s\,{=}\,2$, $v\,{=}\,24$, and $m\,{=}\,6$. The solid curves measure the cosine similarity between the last-layer representation of the original and transformed data, where transformations alter the RHM tree structure. \textbf{Left:} Random replacements of hidden variables at depth $\ell$ indicated by colour, always applied to the penultimate hidden variable ($i\,{=}\,-2$). \textbf{Right:} Random replacements of production rules emanating from the penultimate hidden variable at depth $\ell$. The test loss dynamics are shown as a black dashed line for reference.}
    \label{fig:sensitivities-mla}
\end{figure*}
\autoref{fig:sensitivities-mla} reports the evolution of $q_{\mathcal{T}}$ for the last-layer representation $h_{L}$ of a Transformer during training. The figure shows the last representation, with $i\,{=}\,-1$, but all positions display the same behaviour. The transformations considered are the random replacement of the penultimate observable token, $x_{-2}$, the penultimate hidden variable in each level of the tree, $x^{L-\ell}_{-2}$, and the production rule immediately below. Initially, all the cosine similarities start at different baseline values, reflecting the varying extent to which each transformation alters the input sequence. In correspondence with the first sharp drop of the test loss, all invariance scores sharply increase to $1$, \textit{except} for the one associated with resampling the penultimate observable token (blue line on the left panel). This behaviour indicates that, at this stage, the model relies predominantly on the token immediately preceding the masked token for its prediction.

Subsequently, as training progresses, the invariances to resampling deeper hidden symbols drop sequentially, in order of depth. This sequence of drops signals that the model progressively incorporates longer-range dependencies, leveraging increasingly larger portions of the context window. Notably, the invariance to \textit{symbol replacements} drops significantly at each stage, whereas the invariance to \textit{rule replacements} recovers and approaches unity after each transition. This behaviour strongly suggests that the hidden representations encode the high-level hidden variables associated with the hierarchical structure, while becoming increasingly insensitive to the specific realisations of the subtrees they generate.

\begin{figure*}
    \centering
    \includegraphics[width=\linewidth]{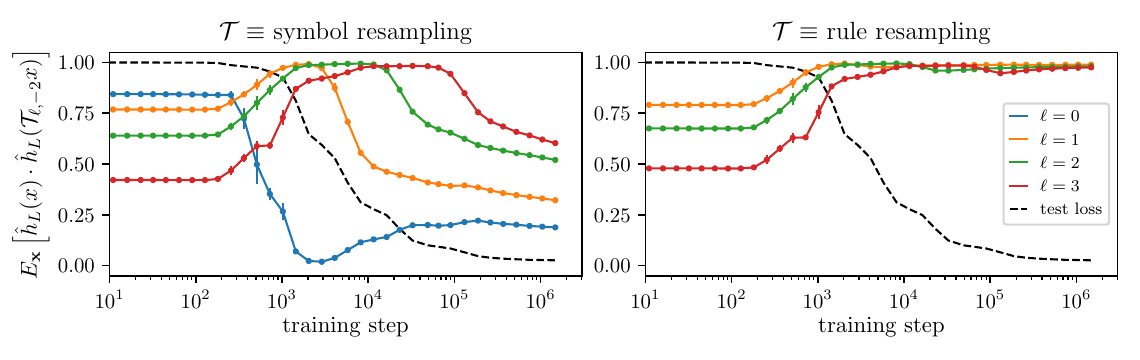}
    \caption{Dynamics of hidden representations for a depth-$4$ CNN trained on RHM data with $L\,{=}\,4$, $s\,{=}\,2$, $v\,{=}\,32$, and $m\,{=}\,8$. The figure follows the layout of~\autoref{fig:sensitivities-mla}.}
    \label{fig:sensitivities-hcnn}
\end{figure*}

As shown in~\autoref{fig:sensitivities-hcnn}, the hidden representations of hierarchical CNNs display the same phenomenology.

\section{Locality Without Weight-Sharing: Locally Connected Networks}\label{sec:lcn}

In this section, we extend our analysis to Locally Connected Networks (LCNs). These architectures retain the local connectivity of CNNs but remove weight sharing across spatial locations. By comparing their performance to that of CNNs and Transformers, we aim to isolate the role of weight sharing in achieving the improved scaling of the test loss displayed by CNNs.

\begin{figure*}
    \centering
    \includegraphics[width=\linewidth]{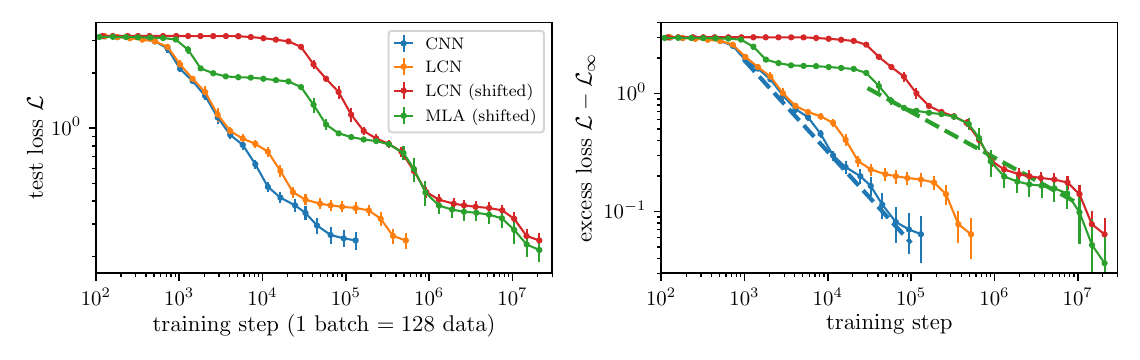}
    \caption{Comparison of training dynamics of depth-$4$ CNNs, Transformers, and LCNs on RHM data with $L=4$, $s=2$, $v=24$, and $m=6$.textbf{Left:} Test cross-entropy loss as a function of the number of training steps. The LCN loss curve (orange) follows that of CNNs (blue) for the first two loss steps, but deviates afterwards. Rescaling the LCN curve by multiplying the $x$-axis by a constant (red curve) reveals that the later steps overlap with those of the Transformer loss curve (green curve, also shifted for visualisation purposes). Right: Scaling of the excess test loss, $\mathcal{L} - \mathcal{L}_{\infty}$, confirming the shared asymptotic behaviour after the shift.}
    \label{fig:hlcn-scaling}
\end{figure*}

\autoref{fig:hlcn-scaling} shows the evolution of the test loss of LCNs as a function of training steps, alongside those of CNNs and Transformers. Initially, the LCN loss follows the CNN loss closely, with the first two steps---corresponding to the inference of depth-$1$ hidden variables---occurring at similar times. However, from the third step onward, the LCN loss deviates: the number of training steps required for the third loss step is noticeably larger than for CNNs, and the fourth step requires even more training. The shape of the loss curve beyond the second step closely matches that of the Transformer, up to a constant rescaling of the number of training steps. This observation suggests that the improved scaling of CNNs at later stages is primarily due to their weight-sharing structure.

% \autoref{fig:sensitivity-hlcn} examines the scaling of relative sensitivities in LCNs, following the same methodology applied to Transformers and CNNs. The relative sensitivity curves associated with depth-1 and depth-2 hidden variables are compatible with the scaling predicted for CNNs. However, the curves corresponding to depth-3 hidden variables deviate from this scaling. A definitive confirmation of this trend would require studying architectures trained on RHM datasets with larger depth $L$, which is currently beyond our computational resources.
% \begin{figure*}
%     \centering
%     \includegraphics[width=\linewidth]{sensitivity-scaling_L4_s2_hlcn_d4_comp-v.pdf}
%     \caption{Relative sensitivity of depth-$(\ell+1)$ representations of hierarchical LCNs trained on RHM data ($L\,{=}\,4$, $s\,{=}\,2$) to depth-$\ell$ transformations, following the layout of~\autoref{fig:sensitivity-scaling-mla}.}
%     \label{fig:sensitivity-hlcn}
% \end{figure*}

\section{Conclusions}\label{sec:conclusions}

In this work, we studied the learning dynamics of deep neural networks trained on hierarchically structured data. Leveraging synthetic datasets generated by the Random Hierarchy Model (RHM)—an analytically tractable ensemble of probabilistic context-free grammars—we analysed how the test loss evolves as a function of training steps under online stochastic gradient descent. Our theoretical framework elucidates how the observed power-law scaling of performance with the number of training steps (or, equivalently, training samples) arises naturally through the progressive acquisition of increasingly deeper layers of the underlying hierarchical structure.

By comparing convolutional networks, transformers, and locally connected networks, we demonstrated that different architectural priors lead to systematic differences in learning dynamics and scaling laws. In particular, convolutional networks tailored to the hierarchical structure of the data achieve a faster scaling of performance. This finding seems to contradict the common intuition that Transformer-based models are better suited for sequence-modelling tasks. However, this contrast highlights the opportunity to use controlled synthetic datasets as probes to uncover the conditions under which different inductive biases achieve better performances. In our case, the strict hierarchical structure of the data aligns naturally with the design of convolutional networks, giving them the edge. One natural extension is to consider variable tree topologies, introducing spatial heterogeneities that might be better exploited by the flexibility of attention mechanisms. Another is to incorporate context-sensitive dependencies, pushing beyond the context-free structure. Exploring such modifications offers a systematic path to understanding when and how Transformer architectures gain their empirical edge.

Several extensions are also possible within the framework of the current data model. On the theoretical side, a formal characterisation of the training dynamics of deep networks on hierarchical data could sharpen our understanding of the mechanisms driving representation learning and the ensuing scaling behaviour. The empirical analysis of internal representations presented in~\autoref{sec:dynamics} provides a promising starting point, as it identifies the information encoded at different layers of trained networks and how this evolves with training. On the empirical side, applying similar analyses to real-world datasets with known hierarchical structure would test the generality of our framework and could guide the design of architectures better suited for compositional learning in practice.

\begin{acknowledgments}
The work of FC was supported by the European Union’s Horizon Europe program under the Marie Skłodowska-Curie grant agreement No. 101154584.
\end{acknowledgments}

\bibliography{PRE2025}% Produces the bibliography via BibTeX.

\end{document}

%% file: parse_tree.tex
% \documentclass{standalone}
% \usepackage{tikz}
% \usetikzlibrary{positioning, arrows.meta}
% \begin{document}
\begin{tikzpicture}[
    var/.style={circle, draw=black, thick, minimum size=7mm, inner sep=1pt},
    term/.style={draw=none, inner sep=1pt},
]

\tikzset{
  mytriangle/.tip = {Triangle[angle=60:2mm]},
  arrowdown/.style = {->, >={mytriangle}, line width=0.25mm}
}

% Terminal nodes (bottom row)
\node[term] (The)     at (0, -1.4) {The};
\node[term] (cat)     at (1.8, -1.4) {cat};
\node[term] (that)    at (3.3, -2.8) {that};
\node[term] (the)     at (4.5, -4.2) {chased};
\node[term] (mouse)   at (5.7, -4.2) {the};
\node[term] (chased)  at (6.9, -4.2) {mouse};
\node[term] (sleeps)  at (7.2, -1.4) {sleeps};

% Intermediate nonterminals
\node[var] (Det)   at (0, 0) {Det};
\node[var] (N)     at (1.8, 0) {N};
\node[var] (C)     at (4.5, 0) {CP};
\node[var] (V)     at (7.2, 0) {V};

% First branch below C: relative clause C and S
\node[var] (C2)    at (3.3, -1.4) {C};
\node[var] (Srel)  at (5.7, -1.4) {S};

\node[var] (Det2)  at (4.5, -4.2 + 1.4) {V};
\node[var] (N2)    at (5.7, -4.2 + 1.4) {Det};
\node[var] (V2)    at (6.9, -4.2 + 1.4) {N};

% Mid-level nodes
\node[var] (NP) at (2, 1.4) {NP};
\node[var] (VP) at (7.2, 1.4) {VP};
\node[var] (S)  at (4.5, 2.8) {S};

% Arrows: S → NP and VP
\draw[arrowdown] (S.south) -- (NP.north);
\draw[arrowdown] (S.south) -- (VP.north);

% Arrows: NP → Det, N, C
\draw[arrowdown] (NP.south) -- (Det.north);
\draw[arrowdown] (NP.south) -- (N.north);
\draw[arrowdown] (NP.south) -- (C.north);

% Arrows: VP → V
\draw[arrowdown] (VP.south) -- (V.north);

% Arrows to terminals
\draw[arrowdown] (Det.south) -- (The.north);
\draw[arrowdown] (N.south) -- (cat.north);
\draw[arrowdown] (V.south) -- (sleeps.north);

% Arrows: C → C2 and Srel
\draw[arrowdown] (C.south) -- (C2.north);
\draw[arrowdown] (C.south) -- (Srel.north);

% Arrow: C2 → "that"
\draw[arrowdown] (C2.south) -- (that.north);

% Arrows: Srel → NP2 and VP2
\draw[arrowdown] (Srel.south) -- (Det2.north);
\draw[arrowdown] (Srel.south) -- (N2.north);
\draw[arrowdown] (Srel.south) -- (V2.north);

% Terminals under Srel
\draw[arrowdown] (Det2.south) -- (the.north);
\draw[arrowdown] (N2.south) -- (mouse.north);
\draw[arrowdown] (V2.south) -- (chased.north);

\end{tikzpicture}
% \end{document}

%% file: rhm_tree.tex
\begin{tikzpicture}[
    varnode/.style={circle, draw=black, thick, minimum size=7mm, inner sep=0pt},
    fac/.style={rectangle, draw, thick, fill=black, minimum size=2mm},
    plain/.style={line width=0.25mm}
]

\tikzset{
  mytriangle/.tip = {Triangle[angle=60:2mm]},
  arrowdown/.style = {->, >={mytriangle}, line width=0.25mm}
}

% Labels
\newcommand{\getlabel}[2]{%
    \ifcase#1%
        \ifcase#2 $g$\or $h$\or $i$\or $j$\or $g$\or $h$\or $k$\or $l$\else $a$\fi
    \or
        \ifcase#2 $d$\or $e$\or $d$\or $f$\else $b$\fi
    \or
        \ifcase#2 $b$\or $c$\else $c$\fi
    \else
        $a$
    \fi
}

% Layout settings
\newcommand\layers{3}
\newcommand\initialdistance{.25cm}
\newcommand\laySpace{.06cm}
\newcommand\leafOffset{0.8} % match offset to factor nodes

%%% Create nodes
\foreach \layer in {\layers,...,0} {
    \pgfmathsetmacro\offX{\initialdistance/(2^(\layer+1))};
    \pgfmathsetmacro\Xdistance{\initialdistance/(2^\layer)};
    \pgfmathsetmacro\numNodes{2^\layer-1};
    \foreach \n in {0,...,\numNodes} {
        \pgfmathsetmacro\Xposition{\offX + \Xdistance * \n};
        \pgfmathsetmacro\Yposition{-\laySpace * \layer};
        \pgfmathtruncatemacro\LayerIndex{\layers-\layer}
        \node [varnode] (\layer-\n) at (\Xposition, \Yposition) {\getlabel{\LayerIndex}{\n}};
        \ifnum\layer<\layers
            \node[fac] (f\layer-\n) at (\Xposition, \Yposition-0.8) {};
        \fi
    }
}

%%% Draw connections
\foreach \layer in {\layers,...,0} {
    \pgfmathsetmacro\numNodes{2^\layer-1};
    \foreach \n in {0,...,\numNodes} {
        \pgfmathtruncatemacro\prevLay{\layer-1};
        \pgfmathtruncatemacro\uppNode{int(\n/2)};
        \ifnum\layer>0
            \draw[arrowdown] (f\prevLay-\uppNode) -- (\layer-\n);
        \fi
        \ifnum\layer<\layers
            \draw[plain] (\layer-\n) -- (f\layer-\n);
        \fi
    }
}

%%% Add horizontal axis directly below leaves, same offset as to factor nodes
\pgfmathsetmacro\leafLayer{\layers}
\pgfmathtruncatemacro\nLeaves{2^\leafLayer}
\pgfmathsetmacro\xStart{\initialdistance/(2^(\leafLayer+1))}
\pgfmathsetmacro\xEnd{\initialdistance - \xStart}
\pgfmathsetmacro\yAxis{-\laySpace * \leafLayer - 0.6}

% Draw axis line
\draw[->, thick] (\xStart, \yAxis) -- (\xEnd + 0.3, \yAxis)
    node[right] {}
    node[midway, below=7pt] {(Position)};

% Add tick marks and labels (one per leaf)
\foreach \n [evaluate=\n as \label using int(-\nLeaves + \n)] in {0,...,7} {
    \pgfmathsetmacro\x{\xStart + (\initialdistance/(2^\leafLayer)) * \n}
    \draw[thick] (\x, \yAxis+0.03) -- (\x, \yAxis-0.03);
    \node[scale=0.75] at (\x, \yAxis - 0.15) {\label};
}

% Compute constants for axis positions
\pgfmathsetmacro\axisLeftX{\xStart - 0.6}
\pgfmathsetmacro\axisRightX{\xEnd + 0.55}
\pgfmathsetmacro\axisBottomY{-\laySpace * 3}  % leaf level
\pgfmathsetmacro\axisTopY{0.0}
\pgfmathsetmacro\axisExtend{0.20} % small extra length                % root level

% Left vertical axis (Depth, up from leaves)
\draw[->, thick] 
    (\axisLeftX, {\axisBottomY - \axisExtend}) -- 
    (\axisLeftX, {\axisTopY + \axisExtend})
    node[above] {Depth};

\foreach \d in {0,...,3} {
    \pgfmathsetmacro\ytick{\axisBottomY + \laySpace * \d}
    \draw[thick] (\axisLeftX - 0.03, \ytick) -- (\axisLeftX + 0.03, \ytick);
    \node[scale=0.75, anchor=east] at (\axisLeftX - 0.05, \ytick) {\d};
}

% Right vertical axis (Level, down from root, title at top)
\draw[->, thick] 
    (\axisRightX, {\axisTopY + \axisExtend}) -- 
    (\axisRightX, {\axisBottomY - \axisExtend});

% Manually add axis title above the top
\node[above] at (\axisRightX, {\axisTopY + \axisExtend}) {Level};
    node[below] {Level};

\foreach \l in {0,...,3} {
    \pgfmathsetmacro\ytick{\axisTopY - \laySpace * \l}
    \draw[thick] (\axisRightX - 0.03, \ytick) -- (\axisRightX + 0.03, \ytick);
    \node[scale=0.75, anchor=west] at (\axisRightX + 0.05, \ytick) {\l};
}

\end{tikzpicture}